\documentclass[conference]{IEEEtran}
\usepackage[utf8]{inputenc}
\usepackage{preamble}
\usepackage[noadjust]{cite}
\usepackage{url}

\title{Finding Streams in Knowledge Graphs\\to Support Fact Checking}

\author{
\IEEEauthorblockN{Prashant Shiralkar\IEEEauthorrefmark{1}, Alessandro Flammini\IEEEauthorrefmark{1}\IEEEauthorrefmark{2}, Filippo Menczer\IEEEauthorrefmark{1}\IEEEauthorrefmark{2}, Giovanni Luca Ciampaglia\IEEEauthorrefmark{2}}
\IEEEauthorblockA{\IEEEauthorrefmark{1}Center for Complex Networks and Systems Research, School of Informatics and Computing}
\IEEEauthorblockA{\IEEEauthorrefmark{2}Network Science Institute\\
Indiana University, Bloomington (USA)
}
}

\begin{document}

\maketitle

\begin{abstract}
    The volume and velocity of information that gets generated online limits current journalistic practices to fact-check claims at the same rate. Computational approaches for fact checking may be the key to help mitigate the risks of massive misinformation spread. Such approaches can be designed to not only be scalable and effective at assessing veracity of dubious claims, but also to boost a human fact checker's productivity by surfacing relevant facts and patterns to aid their analysis. To this end, we present a novel, unsupervised network-flow based approach to determine the truthfulness of a statement of fact expressed in the form of a {\tt (subject, predicate, object)} triple. We view a knowledge graph of background information about real-world entities as a flow network, and knowledge as a fluid, abstract commodity. We show that computational fact checking of such a triple then amounts to finding a ``knowledge stream'' that emanates from the {\tt subject} node and flows toward the {\tt object} node through paths connecting them. Evaluation on a range of real-world and hand-crafted datasets of facts related to entertainment, business, sports, geography and more reveals that this network-flow model can be very effective in discerning true statements from false ones,  outperforming existing algorithms on many test cases. Moreover, the model is expressive in its ability to automatically discover several useful path patterns and surface relevant facts that may help a human fact checker corroborate or refute a claim. 
\end{abstract}


\section{Introduction}
\label{sec:introduction}

Misinformation, unverified rumors, hoaxes, and lies have become rampant on the Internet nowadays, primarily due to the ability to quickly disseminate information at a large scale through the Web and social media. This phenomenon has led to many ill effects and, according to experts, poses a severe threat to society at large~\cite{Howell2013}. To address these problems, numerous approaches have been designed to study and mitigate the effects of misinformation spread (see Zubiaga~\textit{et al.}~\cite{zubiaga2017detection}). Most strategies rely on contextual indicators of rumors (e.g., number of inquiring tweets, reporting dynamics during breaking news, temporal patterns, or source credibility) for their detection and veracity assessment. 
To go beyond contextual approaches one would need to assess the truthfulness of claims by reasoning about their content and related facts. Moreover, a fact-checking system would ideally need to operate in near real time, to match the rate at which false or misleading claims are made.
    
\begin{figure}[tbp]
    \centering
    \includegraphics[scale=0.4]{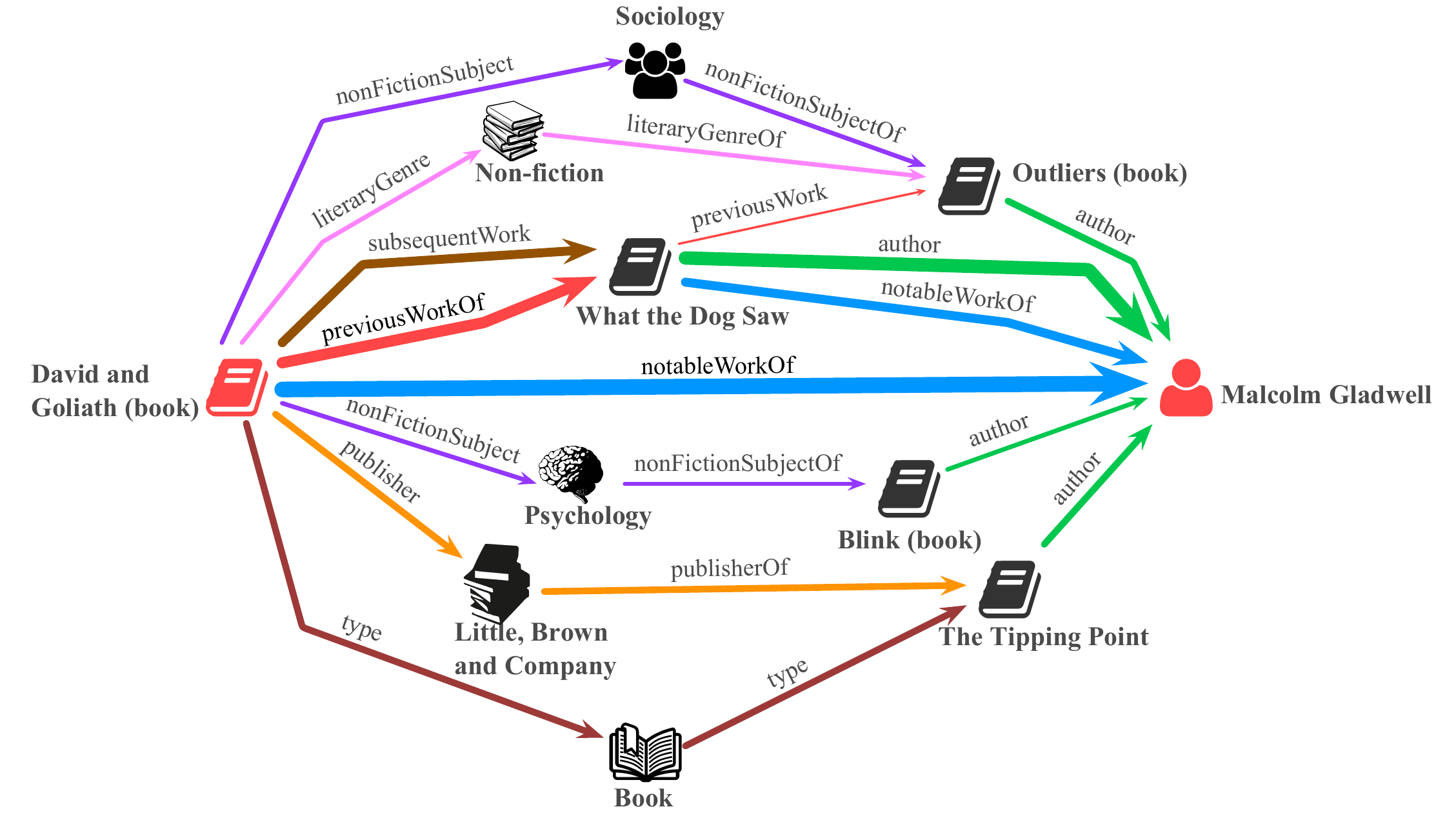}
    \caption{The best paths identified by Knowledge Stream for the triple {\tt (David and Goliath (book), author, Malcolm Gladwell)}. The width of an edge is roughly proportional to the flow of knowledge through it.}
    \label{fig:ks-example}
\end{figure}

With advances in information extraction and in the adoption of semantic web standards, large quantities of structured knowledge have recently become available in the form of knowledge graphs (KGs). Nodes in a KG represent entities, and edges correspond to facts about them, as specified by semantic predicates, or relations. A wide class of empirical facts can be thus represented by a triple $(\triple)$, where the subject entity $s$ is related to the object entity $o$ by the predicate relation $p$. For example, {\tt (Joe, spouse, Jane)} indicates that Jane is the spouse of Joe. DBpedia~\cite{bizer2009dbpedia}, YAGO2~\cite{hoffart2013yago2} and Wikidata~\cite{erxleben2014introducing} are examples of publicly available KGs. These KGs contain vast amounts of high-quality knowledge about real-world entities, events, and their relations, and thus could be at least in principle harnessed by fact-checking agents.

Insofar as claims as simple as a triple are of concern, how can we automatically assess their truthfulness, given a large amount of prior knowledge structured as a KG? A few recent attempts have shown that this is possible via traversal of the graph. Traversal can take many forms, for example random walks (PRA~\cite{lao2010relational}), path enumeration (PredPath~\cite{shi2016discriminative}), or shortest paths (Knowledge Linker~\cite{ciampaglia2015computational}). Other approaches have been proposed too, such as those designed for learning from multi-relational data (e.g., RESCAL~\cite{nickel2011three}, TransE~\cite{bordes2013translating} and their extensions), or those performing link prediction in social and collaboration networks~\cite{liben2007link}. 

However, as we discuss in~\prettyref{sec:related-work}, none of these approaches offers at once all the qualities that a desirable fact-checking system ought to have --- namely accuracy, interpretability, simplicity, scalability, and the ability to take the greater context of a claim into account while evaluating it.

In this paper we propose Knowledge Stream (KS), an unsupervised approach for fact-checking triples based on the idea of treating a KG as a flow network.  There are three motivations for this idea: (1) multiple paths may provide greater semantic context than a single path; (2) because the paths are in general non-disjoint, the method reuses edges participating in diverse overlapping chains of relationships by sending additional flow; and (3) the limited capacities of edges limit the number of paths in which they can participate, constraining the path search space.

Our approach not only delivers performance comparable to state-of-the-art methods like PredPath; it also produces more meaningful explanations of its predictions. It does so by automatically discovering in the KG useful patterns and contextual facts in the form of paths. The model we propose is conceptually simple, intuitive, and uses the broader structural and semantic context of the triple under evaluation. 

As an example, \prettyref{fig:ks-example} shows the paths computed for a true fact {\tt (David and Goliath (book), author, Malcolm Gladwell)}. We call this set of paths a ``stream'' of knowledge. A stream can thus be seen as the best form of evidence in support of the triple that the KG is able to offer. One can note from \prettyref{fig:ks-example} that some paths give more evidence than others (wider edges in the figure). For example, the fact that Malcolm Gladwell is the author of the book \textit{What the Dog Saw}, which followed \textit{David and Goliath}, is a stronger form of evidence than the fact that another book authored by Gladwell, \textit{The Tipping Point}, was published by the same company (Little, Brown and Company) as \textit{David and Goliath}. Knowledge Stream correctly assigns a larger flow to the former path than the latter.

For a given triple $(\triple)$, we view knowledge as a certain amount of an abstract commodity that needs to be moved from the subject entity $s$ to the object entity $o$ across the network. Each edge of the network is associated with two quantities: a \emph{capacity} to carry knowledge related to $(\triple)$ across its two endpoints, and a \emph{cost} of usage. We want to identify the set of paths responsible for the maximum flow of knowledge between $s$ and $o$ at the minimum cost. 
We give some definitions to make these statements more formal and explain the intuition behind our approach.

Each edge $e\in E$ of the KG has an intrinsic \emph{capacity}, which depends on the triple under consideration. Recall that an edge is labeled with a predicate $p'$ possibly different from the target predicate $p$. Intuitively, the more similar, or relevant, $p'$ is to $p$, the higher the capacity of $e$ ought to be. If we are to ascertain whether Jane is indeed the spouse of Joe, facts about the realm of, say, geology are in general less pertinent than facts about ancestry or family history. 
We use a data-driven approach, mining the structure of the KG itself, to define the similarity between predicates. To do so, we employ the graph-theoretic concept of \emph{line graph} of the KG. The full details are described in~\prettyref{sec:relational-similarity}.

The maximum knowledge flow carried by a path is the minimum capacity of its edges. The edge at which the minimum is found is known as the \emph{bottleneck} of the path~\cite{ahuja1993network}.
In our approach, the bottleneck corresponds to the least relevant triple along the path. In general, there are many paths connecting $s$ to $o$, and the total knowledge that can flow through them is bounded by the sum of their bottlenecks.

To each edge $e\in E$ we also associate a \emph{cost} for sending a unit of flow across its two endpoints. This ensures that the paths discovered by KS are short. Previous 
work has directly or indirectly confirmed the intuition that the structures (walks, paths, etc.) that best explain whether a triple is true are short~\cite{lao2010relational,ciampaglia2015computational,shi2016discriminative}. 

Our definition of path length differs from the traditional number of hops: a short path involves not only few entities but also entities with few connections to other entities in the graph~\cite{ciampaglia2015computational}. We say that such entities are ``specific,'' and the the paths containing them are ``specific paths.''

As mentioned earlier, one of the components of KS is the method to compute similarity between relations. This method can also be applied to shortest-path approaches, such as Knowledge Linker. The resulting algorithm, which we call ``Relational Knowledge Linker'' (KL-REL), assigns a truth score to $(\triple)$ by biasing the search for the shortest path toward predicates related to $p$. 

In summary, this paper makes the following contributions:
\begin{itemize}
    \item We propose a novel method called Knowledge Stream to perform computational fact checking using large knowledge graphs such as DBpedia~\cite{bizer2009dbpedia}. To our knowledge, this is the first instance of applying flow network to the problem of soft reasoning with knowledge graphs. 
    \item We introduce a novel approach to gauge similarity between a pair of relations in the KG. This similarity can be used for many other tasks, e.g. analogical reasoning.
    \item We propose a fact-checking algorithm called Relational Knowledge Linker that verifies a claim based on the single shortest, semantically related path in the KG.
    \item We experimentally compare our approaches of Knowledge Stream and Relational Knowledge Linker to a number of existing algorithms designed for fact checking, knowledge graph completion, and link prediction. We show that both KS and KL-REL offer high interpretability and performance comparable to the state of the art.
\end{itemize}

\section{Methods}
\label{sec:knowledge-stream}

In this section we describe Knowledge Stream and Relational Knowledge Linker, the two methods we propose to perform fact checking using a KG. Formally, a KG is a directed graph $G = (\nodeset, \edgeset, \relationset, \labelfun)$, where $\nodeset$, $\edgeset$, and $\relationset$ denote the node, edge, and relation sets, respectively, and $\labelfun:\edgeset\rightarrow\relationset$ is a function labeling each edge with a semantic relation or predicate. Even though $G$ is a directed network, in practice most existing methods for fact checking, including ours, view it as an undirected one by discarding the directionality of edges. 

Since both methods presented here rely on the ability to gauge the similarity, or relevance, of any pair of elements of $\relationset$, we start by explaining our data-driven approach to relational similarity. 

\subsection{Relational Similarity via the Line Graph of a KG}
\label{sec:relational-similarity}

\begin{figure}[tbp]
    \centering
    \includegraphics[width=3in]{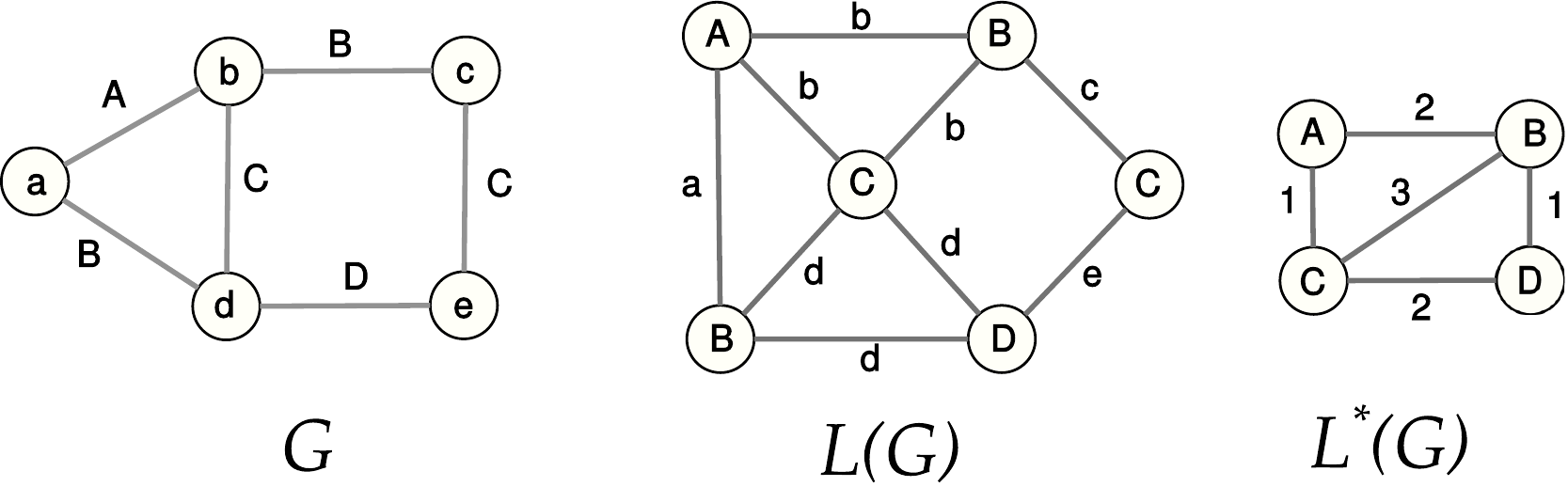}
    \caption{Example of the line graph $L(G)$ and the contracted line graph $L^*(G)$ of a simple knowledge graph $G$ with four relations (denoted by uppercase letters) and five nodes (lowercase letters). The edge weights in $L^*(G)$ represent how often each relation is co-incident to its neighbors in $G$.}
    \label{fig:linegraph}
\end{figure}

In graph theory, the \emph{line graph} $L(G) = (\nodeset', \edgeset')$ of an undirected graph $G=(\nodeset,\edgeset)$ is the graph whose nodes set is $\nodeset'=\edgeset$ and in which two nodes are adjacent \emph{iff} the corresponding edges of $G$ are incident on the same node in $G$, that is, $\edgeset' = \left\{(e_1, e_2): e_1, e_2 \in \edgeset \wedge e_1 \cap e_2 \neq \emptyset \right\}$. 
Line graphs are also sometimes known as \emph{dual graphs}. The KG being an edge-labeled graph makes the $L(G)$ a node-labeled graph. However, even though $L(G)$ encodes information about the adjacency between relations of the KG, it is not suited to define a similarity metric on $\relationset$ because it includes duplicate labels. We overcome this problem by contracting duplicate nodes until there is exactly one node for each element of $\relationset$. A graph can be \emph{contracted} by replacing two nodes with a new node whose set of neighbors is the union of their neighbors. Rather than duplicating edges, the contracted graph is edge-weighted; the weight of a new edge reflects the number of old edges that are merged in the contraction. We thus start from $G$, then build $L(G)$ setting all edge weights to 1, and finally we iteratively contract pairs of nodes labeled with the same relation, until there are no duplicate labels. We call the resulting graph the \emph{contracted line graph}, denoted by $L^*(G)$. See~\prettyref{fig:linegraph} for an example of a small KG with four relations and five nodes.

\begin{figure}[tbp]
    \centering
    \subfloat[spouse]{
        \includegraphics[width=1.6in]{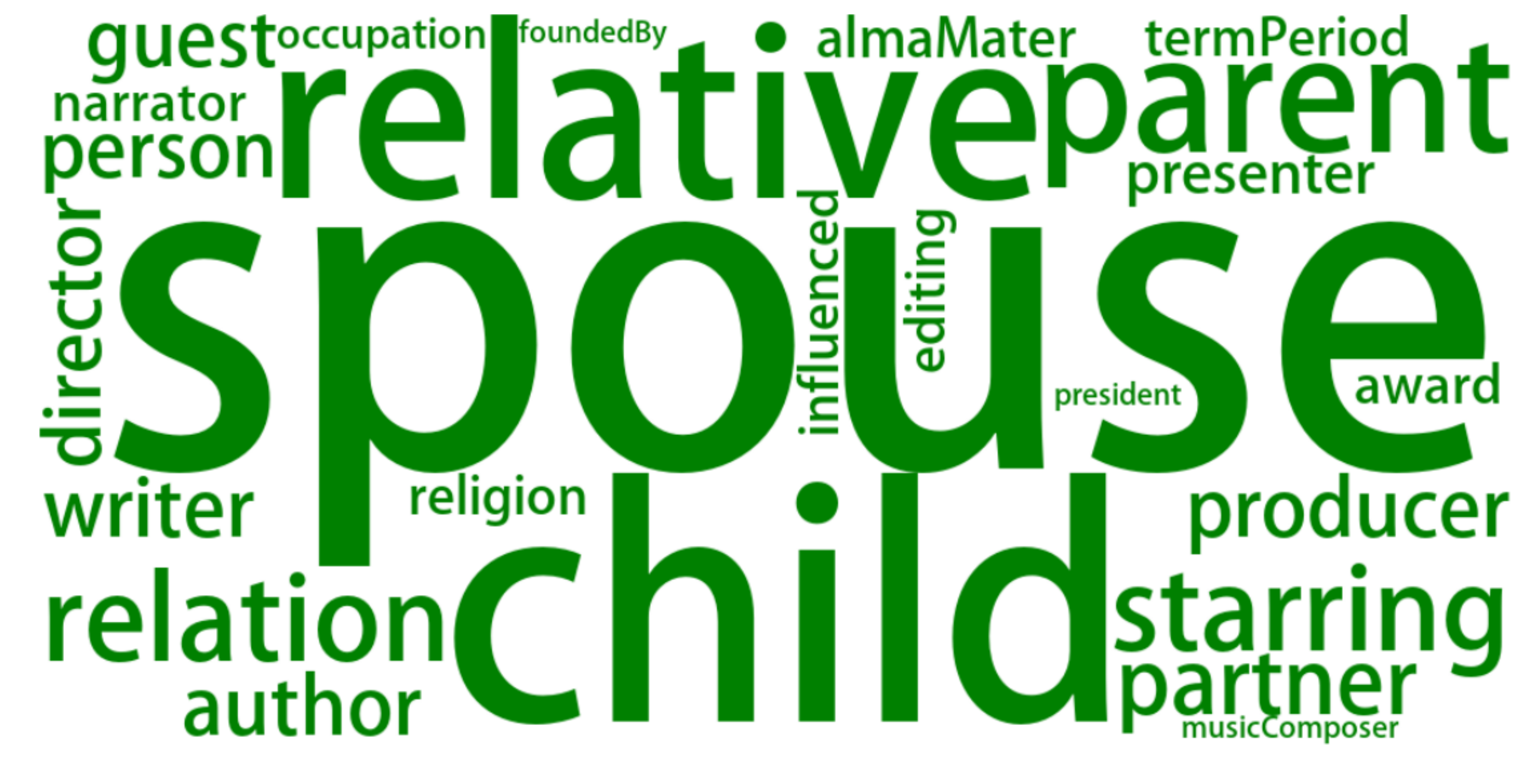}\label{fig:relsim1}
    }
    \hfil
    \subfloat[director]{
        \includegraphics[width=1.6in]{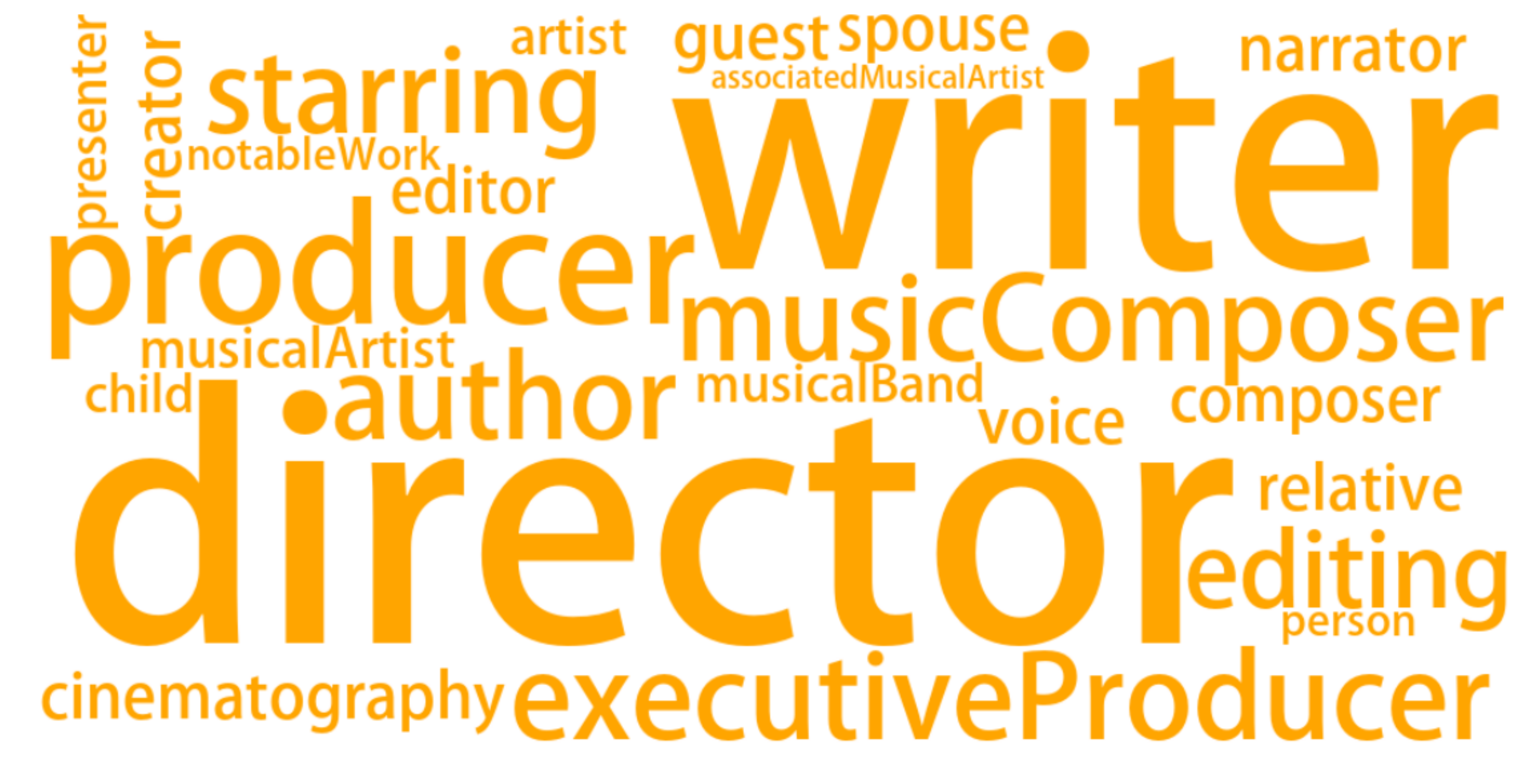}\label{fig:relsim2}
    }
    \hfil
    \subfloat[capital]{
        \includegraphics[width=1.6in]{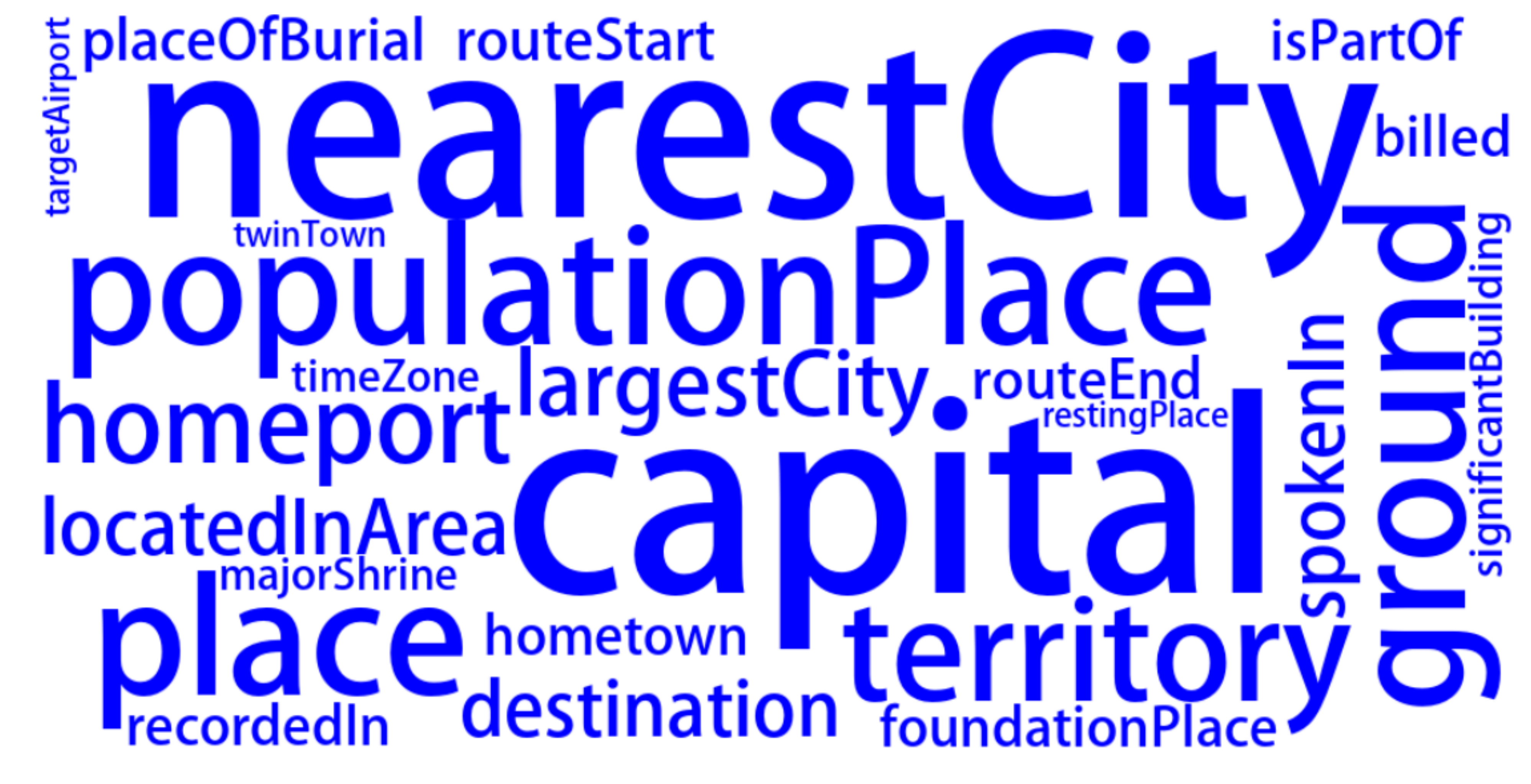}\label{fig:relsim3}
    }
    \hfil
    \subfloat[battle]{
        \includegraphics[width=1.6in]{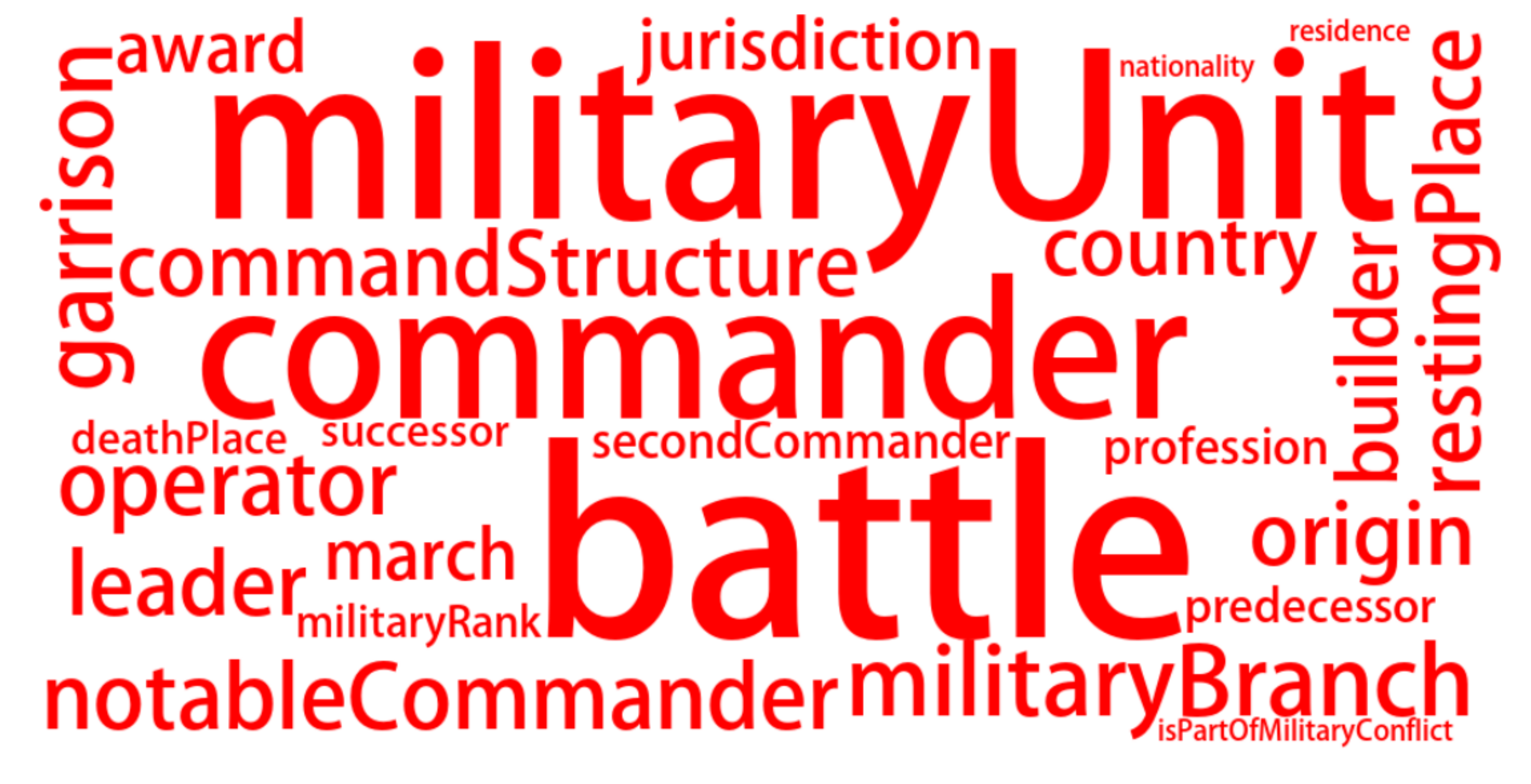}\label{fig:relsim4}
    }
    \hfil
    \subfloat[education]{
        \includegraphics[width=1.6in]{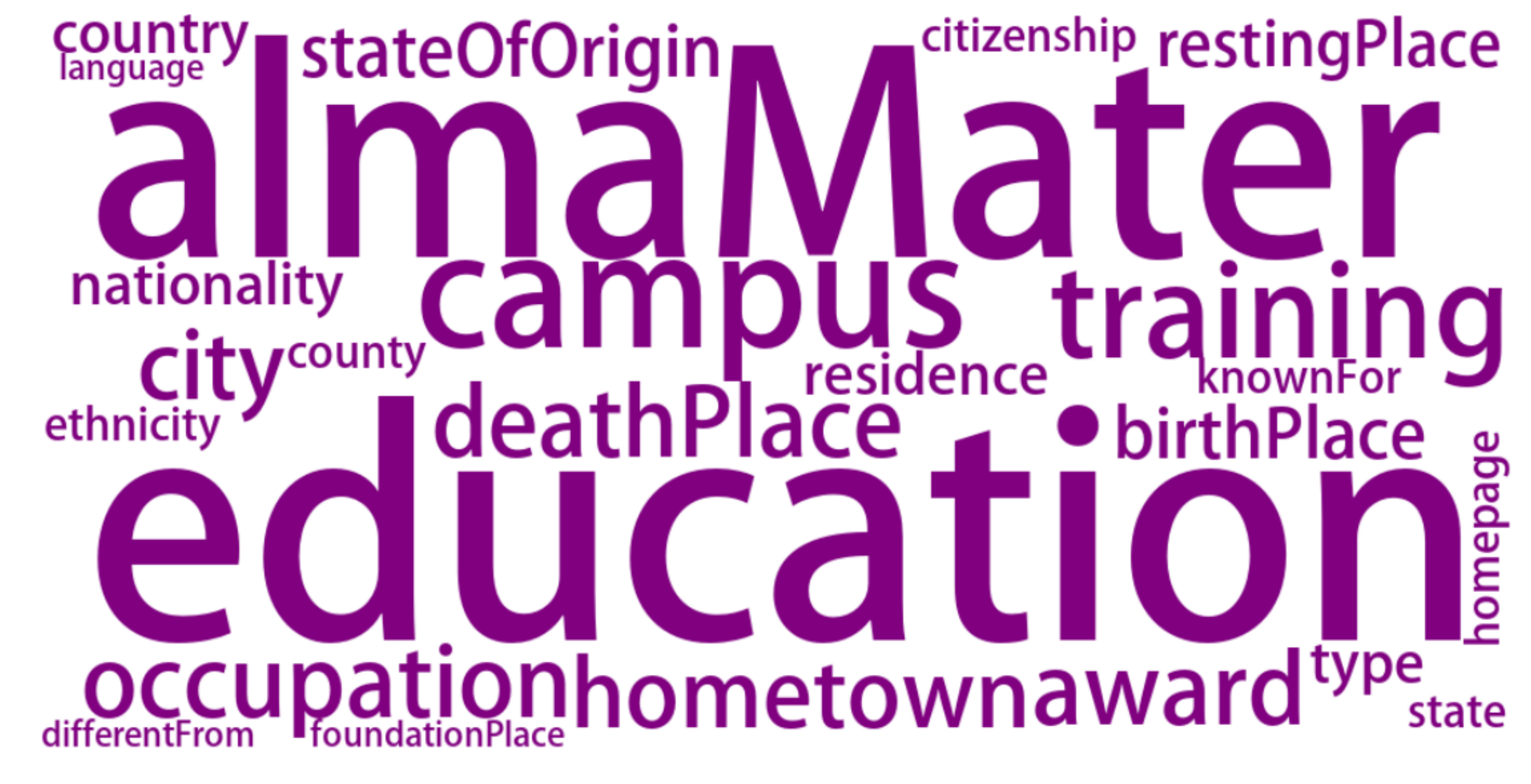}\label{fig:relsim5}
    }
    \hfil
    \subfloat[vicePresident]{
        \includegraphics[width=1.6in]{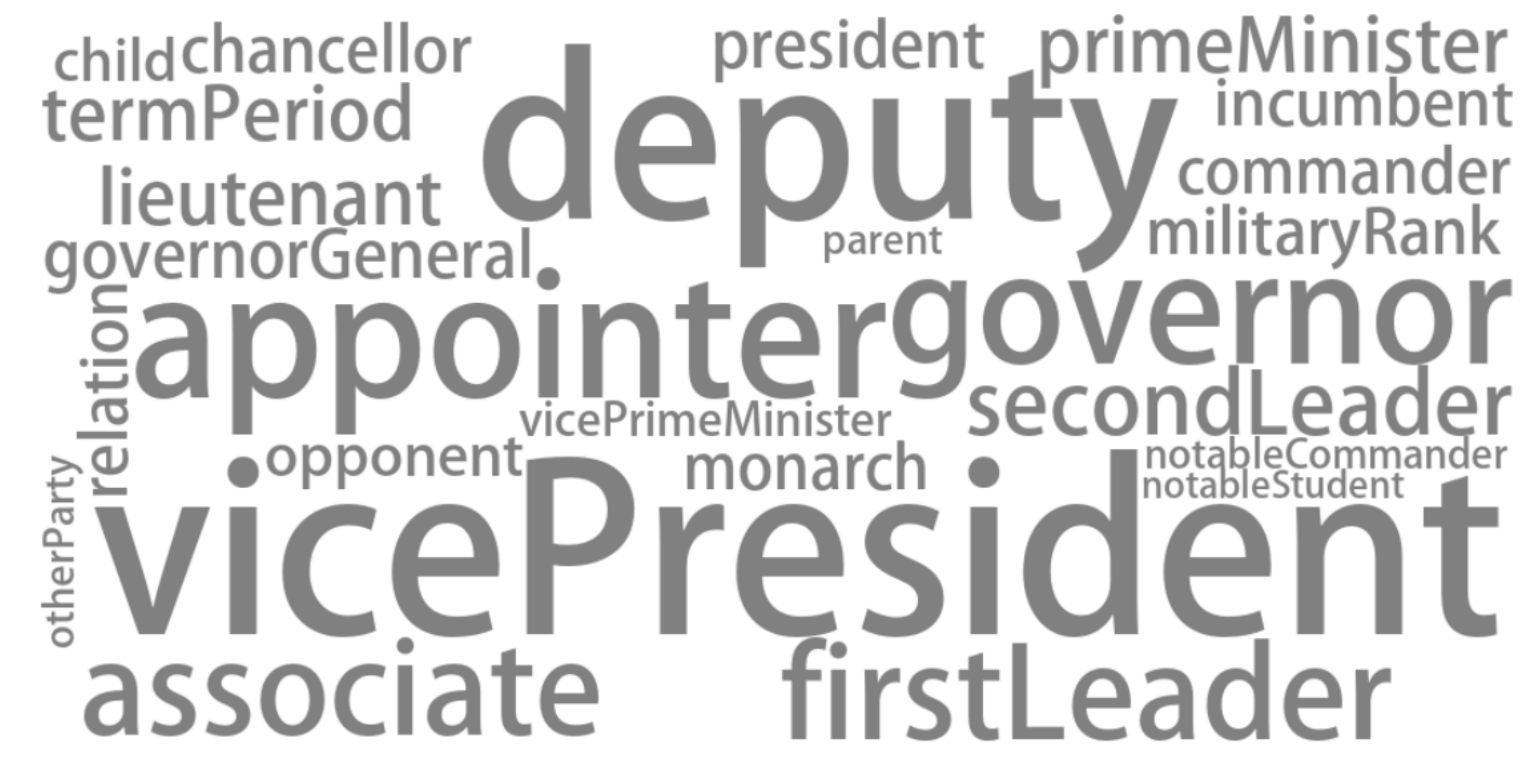}\label{fig:relsim6}
    }
    \caption{Top 20 most similar relations for a few  predicates in DBpedia. The font size is proportional to the relational similarity.}
    \label{fig:relsim-results}
\end{figure}

Let us denote with $C\in \mathbb N^{R\times R}$, where $R=|\relationset|$, the adjacency matrix  of the contracted line graph. By construction, $C$ is the co-occurrence matrix of $\relationset$. One could estimate the similarity between two relationships by computing the cosine between the row vectors corresponding to the relationships in $C$. However, the raw co-occurrence counts in $C$ are dominated by the most common relationships. Therefore, as customary in information retrieval, we apply TF-IDF weighting to $C$: 
\begin{align}
\text{TF}(r_i, r_j) &= \log (1 + C_{ij}), \nonumber \\
\text{IDF}(r_j, \relationset) &= \log \frac{R}{|\{r_i | C_{ij} > 0 \}|}, \nonumber \\
C'(r_i, r_j, \relationset) &= \text{TF}(r_i, r_j) \cdot \text{IDF}(r_j, \relationset) \label{eq:relsim}
\end{align}
where $C_{ij}$ is the co-occurrence count between $r_i\in\relationset$ and $r_j\in\relationset$. 
We define the \emph{relational similarity} $\relsim[r_j]{r_i}$ as the cosine similarity of the $i$-th and $j$-th rows of $C'$. We found that this approach yields meaningful results; a few examples are shown in \prettyref{fig:relsim-results}.

\subsection{Fact checking as a Minimum Cost Maximum Flow Problem}

As stated in the introduction, with Knowledge Stream we view fact checking as the problem of finding an optimal way to transfer, across the KG, knowledge from the source entity to the target entity under a set of constraints. These constraints depend both on the KG itself and on the given relation that we are trying to check. 

The first set of constraints on the edges dictate that the amount of flow that can be pushed across an edge is bounded. In Knowledge Stream, we take the lower bound on this flow to be zero, and we define the upper bound or \emph{capacity} of $e = (v_i, v_j) \in \edgeset$ with respect the triple to be fact-checked, $(\triple)$, as
\begin{equation}
\edgecapacity[(e)] = \frac{u\left(g(e),\, p\right)}{1 + \log k(v_j)},
\label{eq:ks-edgecap}
\end{equation}
which is the product of the similarity $u$ between the edge label $g(e)$ and the predicate $p$ of the target triple (see \prettyref{sec:relational-similarity}), and a quantity that represents the \emph{specificity} of the node to which $e$ is incident. The specificity is based on the assumption that the larger the degree $k$ of a node --- the more facts in the KG about it --- the more \emph{general} the concept is. 
Our use of the logarithm of the degree is based on information-theoretic arguments~\cite{ciampaglia2015computational}. Alternative choices could of course be explored.

The second set of constraints relate to conservation of flow across nodes: except for the nodes corresponding to the subject $s$ and object $o$, the amount of flow entering a node must be equal to that leaving the node. We associate with $s$ (resp. $o$) a fixed \emph{supply} (\emph{demand}) of knowledge, $\streamflow$, which is the maximum feasible flow through the network. 

In network flow problems, costs map to quantities to be minimized, like the distance of a road or amount of gas spent carrying goods over it. In the KG context, we employ again the idea that the degree of a node is a measure of generality, to be minimized. We therefore set the cost of edge $e = (v_i, v_j) \in \edgeset$ to $c_e = \log k(v_j)$. Note that although the KG is undirected, for each edge along a path, the capacity and cost functions consider the degree of the incident node $v_j$ in the direction from $s$ to $o$.

Having defined the main constraints, we solve a \textit{minimum cost maximum flow} problem~\cite[Ch. 1, 9, 10]{ahuja1993network}. The flow assignment to the edges of the KG is a non-negative real-valued mapping $f:\edgeset \rightarrow \mathds{R}^+$, that maximizes the total flow $\streamflow$ pushed from $s$ to $o$ while minimizing the total cost $\sum_{e \in \edgeset}{} c_{e} f(e)$ subject to the edge capacity constraints:
\begin{equation}
    0 \leq f(e) \leq \edgecapacity[(e)] \nonumber
\end{equation}
and to the node conservation constraints:
\begin{align}
     b(v) = 
     \begin{cases} 
        \streamflow & v = s \\ 
        -\streamflow &  v = o \\
        0 & \mbox{otherwise}
    \end{cases} \nonumber
\end{align}
where $b(v_i) = \sum_{v_j \in V} f(v_i, v_j)$ is the net flow outgoing from node $v_i$.

We are interested in finding the set of paths along which the maximum flow $\gamma$ is pushed from $s$ to $o$. In practice we solve the minimum cost maximum flow problem using an algorithm that computes such a set of paths. We denote this set of paths the \emph{stream of knowledge} $\stream$. Each path in the stream carries knowledge at its full capacity. The maximum knowledge a path~$\path$ can carry is the minimum of the capacities of its edges, also called its \textit{bottleneck} $\bottleneck$. It can be shown that the maximum flow is the sum of the bottlenecks of the paths that are part of the stream:
\begin{align}
\streamflow = \sum_{\path \in \stream} \bottleneck.
\label{eq:ks-maxflow}
\end{align}

Having determined the maximum flow and knowing the exact contribution of each individual path in a stream, we need to specify how to use the stream for fact checking. The flow through a path captures the relational similarity and specificity of its bottleneck, as well as the specificity of the intermediate nodes. Nevertheless, long chains of specific relationships could lead us astray. Therefore Knowledge Stream should favor \textit{specific paths} involving few specific entities. We define the specificity~$\specificity$ of a path~$\path$ with $n$ nodes as inversely proportional to the sum of logarithms of the degrees of its intermediate nodes:
\begin{align}
\specificity = \frac{1}{1 + \sum\limits_{i=2}^{n-1}\log k(v_i)}. \label{eq:path-specificity}
\end{align}

We say that the net flow~$\pathweight[]$ in a path~$\path$ is the product of its bottleneck~$\bottleneck$ and specificity~$\specificity$: 
\begin{align}
    \pathweight[] &= \bottleneck \cdot \specificity. 
\end{align} 
Fact checking a triple $(\triple)$ then reduces to computing a \emph{truth score}~$\streamweight[KS]$ as the sum of the net flow across all paths in the stream:
\begin{align}
\streamweight[KS] &= \sum\limits_{\path \in \stream}{} \pathweight[] \nonumber \\
&= \sum\limits_{\path \in \stream}^{} \bottleneck \cdot \specificity. \label{eq:ks-truthval}
\end{align}

\subsection{Computing the Knowledge Stream}
\label{sec:computing-knowledge-stream}

Let us now discuss how to solve our optimization problem and compute the truth score of a triple in practice. A well-known algorithm called  Successive Shortest Path (SSP) provides a solution to the optimization problem and also returns the sequence of paths. The idea is to push the maximum flow~$\streamflow$ from $s$ to $o$ by iteratively finding a shortest path in a residual network, along which we can push some flow.  The \emph{residual network} $G(f)$ of $G$ w.r.t flow $f$ has the same set of nodes $\nodeset$ as $G$, but has two kinds of edges: (1)~\textit{forward edges} with some ``leftover capacity'' over which one can push additional flow, and (2)~\textit{backward edges} that represents edges already allocated, over which one can push reverse flow in order to undo flow in forward edges. At each step in the iteration we compute the bottleneck of the shortest path, given by
\begin{align}
\bottleneck = \min \left\{x_{e} \middle| e \in \path \right\}, \label{eq:ks-bnck}
\end{align}
where $x_{e} \leq \mathcal{U}_{e}$ represents the residual capacity of edge $e$ in the residual network. 
Our extended version of SSP to compute the stream of knowledge and the truth score $\streamweight[KS]$ for a given triple is shown in Algorithm~\ref{alg:knowledge-stream}. 

\begin{algorithm}
\caption{Knowledge Stream Algorithm}
\label{alg:knowledge-stream}
\begin{algorithmic}[1]
\Procedure{KnowledgeStream}{$G$, $\triple$}
    \State $\streamweightsymbol \gets 0, \mathcal{P} \gets \emptyset, f \gets 0$
    \State $\pi \gets 0$ \label{algline:node-potentials}
    \State $c_{\edge} = \log(v_j), \forall (\edge) \in \edgeset$ \label{algline:edge-len}
    \State $c^\pi_{\edge} = c_{\edge} - \pi(v_i) + \pi(v_j)$ \label{algline:reduced-edge-len}
    \State $d \gets \text{compute shortest path distances from $s$ to all}$
    \StatexIndent{other nodes in $G(f)$ w. r. t. $c^\pi$}
    \State $P \gets \text{a shortest path from $s$ to $o$ in $G(f)$}$
    \While {$P$ exists}
        \State $\mathcal{P} \gets \mathcal{P} \cup \{P\}$
        \State $\pi \gets \pi - d$
        \State $\beta(P) \gets \min \left\{x_{\edge} \middle| (\edge) \in P\right\}$
        \State $\text{Push $\beta(P)$ units of flow along $P$}$
        \State $\mathcal{S}(P) \gets \frac{1}{1 + \sum_{i=2}^{n-1}\log k(v_i)} \text{ for $v_i \in P$}$
        \State $\mathcal{W}(P) \gets \beta(P) \cdot \mathcal{S}(P)$
        \State $\streamweightsymbol \gets \streamweightsymbol + \mathcal{W}(P)$
        \State {update $f, G(f)$ and reduced edge lengths $c^\pi$}
        \State $d \gets \text{compute shortest path distances from $s$ to}$ \label{algline:pathfinding}
        \StatexIndent{all other nodes in $G(f)$ w. r. t. $c^\pi$} 
        \State $P \gets \text{a shortest path from $s$ to $o$ in $G(f)$}$ \label{algline:pathfinding2} 
    \EndWhile
    \State \textbf{return} $\streamweightsymbol, \mathcal{P}$
\EndProcedure
\end{algorithmic}
\end{algorithm}

We associate a real number~$\pi(v_i)$ (\prettyref{algline:node-potentials}) with each node $v_i \in \nodeset$, called its \textit{node potential}. A vector of such node potentials~$\pi$ serves two important purposes: (1)~it allows us to keep track of the \textit{reduced cost} $c^{\pi}$~(\prettyref{algline:reduced-edge-len}) of an edge at each step of the algorithm, which makes successive path-finding efficient; and (2)~it serves as an ingredient of the \textit{reduced cost optimality conditions} that ensure the achievement of maximum flow upon termination~\cite[Ch. 9]{ahuja1993network}.

The complexity bounds for the SSP algorithm assume that all edge weights are integral, which does not hold for our capacities ($\edgecapacity[] \in [0,1]$). This is not a problem however, since capacities are rational numbers and can therefore be easily converted to integers.  

If the maximum flow~$\streamflow$ is an integer, the Knowledge Stream algorithm takes at most $\streamflow$ iterations. Since each shortest path computation can be performed in $O(|\edgeset| \log |\nodeset|)$ time using Dijkstra's algorithm~\cite{dijkstra1959note} with a binary heap implementation, the overall complexity of the algorithm is $O(\streamflow |\edgeset| \log |\nodeset|)$. In practice, $\streamflow$ is not an integer, and is computed by the algorithm; this makes Knowledge Stream a pseudo-polynomial time algorithm. In practice we find acceptable performance: for large-scale KGs such as DBpedia, our implementation takes an average of 356 seconds per triple on a laptop. 

\subsection{Relational Knowledge Linker}
\label{sec:rel-knowledge-linker}

Our measure of relational similarity defined in \prettyref{sec:relational-similarity} can also be used to extend existing KG-based fact-checking methods. One such method is Knowledge Linker (KL)~\cite{ciampaglia2015computational}. The approach used by KL for fact checking a triple $(\triple)$ is to find the path between entities $s$ and $o$ that maximizes specificity (\prettyref{eq:path-specificity}). This approach ignores the semantics of the target predicate $p$. 
We hypothesize that biasing the search for specific paths to favor edges that are semantically related to $p$ should improve KL.  
We therefore replace the definition of path specificity in \prettyref{eq:path-specificity} by 
\begin{align}
\mathcal{S}'(\path) &= \left[ 
\sum_{i=2}^{n-1} \frac{\log k(v_i)}{\relsim{r_{i-1}}} +
\frac{1}{\relsim{r_{n-1}}} 
\right]^{-1}. \label{eq:relkl-pathweight}    
\end{align}
This formulation maximizes the relational similarity between each edge and the target predicate, in addition to the specificity of the intermediate nodes. The last term allows to consider the relation of the last edge without penalizing the generality of the object $o$. 
The truth score of triple $(\triple)$ is just 
$\streamweight[KL-REL] = \max_{\path \in \stream} \mathcal{S}'(\path)$.

The truth score and the associated path can be computed efficiently using Dijkstra's~ algorithm~\cite{dijkstra1959note}. We call this extended approach the Relational Knowledge Linker (KL-REL). 

\section{Evaluation}
\label{sec:evaluation}

In this section we present the results of an evaluation of our two methods, Knowledge Stream (KS) and Relational Knowledge Linker (KL-REL), on a range of datasets. To make the evaluation meaningful, we pit these algorithms against a number of existing approaches from the literature on computational fact checking and related problems, namely automated knowledge base construction (KBC) and link prediction. 
We start by describing the experimental setup.


\subsection{Setup}
\label{sec:setup}

\subsubsection{Knowledge Graph} We select DBpedia, a popular knowledge base derived from Wikipedia, as the KG for all evaluations. DBpedia is a large community effort with the goal of extracting structured data from the body and infobox of each Wikipedia article. It is freely available in a serialized form, split across a number of RDF data dumps. In particular, to build the KG we used in the evaluation, we downloaded and merged together the following dumps: ontology, instance-types, and mapping-based properties. We use the most recent distribution at the time of evaluation.\footnote{\url{wiki.dbpedia.org/downloads-2016-04}} We apply the following filtering to the dumps: (1)~from the instance-types dump we remove all subsumption triples (i.e., triples of the form $x~\relation{is a}~T$, where $x$ is an entity and $T$ is a class in the DBpedia ontology) that are the result of transitive closure, as they shortcut the ontological hierarchy in an undesired way; and (2)~we discard all triples whose object is an RDF literal (e.g., dates, numerical values, text labels), as they do not correspond to any KG entity. The undirected graph we obtain as a result has the following characteristics: $|\nodeset| = 6\textrm{M}$ nodes, $|\edgeset|=24\textrm{M}$ triples, and $|\relationset| = 663$ relations.

\subsubsection{Labeled Datasets}
\label{sec:test-cases}

We evaluate all methods on two classes of datasets. The first class includes \emph{synthetic} corpora that have been created for evaluation purposes by our team and others. These datasets mix \textit{a priori} known true and false facts and are drawn from the domains of entertainment, business, geography, literature, sports, etc. Additional datasets include triples extracted in the wild, whose ground truth covers the full spectrum of truth scores, ranging from completely true to completely false. Several \emph{real-world} datasets in the second class are derived from the Google Relation Extraction Corpora (GREC),\footnote{\url{research.googleblog.com/2013/04/50000-lessons-on-how-to-read-relation.html}} which contain information about birth place, death place, alma mater, and educational degree of notable people. Two more datasets about professions and nationalities are derived from the corpus of the WSDM Cup 2017 Triple Scoring challenge.\footnote{\url{www.wsdm-cup-2017.org/triple-scoring.html}} 
\prettyref{tab:datasets} summarizes all the datasets. Those marked with an asterisk were first used in prior work~\cite{shi2016discriminative}. We report the average number of facts per subject in the last column.

The ground truth in both the GREC and the WSDM Cup corpora was obtained via crowdsourcing. In the GREC, each triple was evaluated by five human raters. We use only triples whose rating was unanimous, i.e., either all true or all false. In the WSDM Cup corpus each triple was scored by seven raters, but the corpus contains only true triples, by design. We consider only true triples with a unanimous score, and we generate false facts by randomly drawing from professions or nationalities that individuals are not known to hold. This approach amounts to making a \emph{local closed-world assumption} (LCWA). 

\begin{table*}[tbp]
\renewcommand{\arraystretch}{1.3}
\caption{Summary of datasets used in the evaluation.}
\label{tab:datasets}
\centering
\begin{tabular}{lllcc}  
\toprule
& \textbf{Dataset} & \textbf{Example Fact} & \textbf{True / Total} & \textbf{Facts / Subject} \\
\midrule
\multirow{8}{*}{Synthetic} &
NYT-Bestseller*     & Flash Boys, author, M. Lewis      & $93/558$ & $7.5$ \\ 
& NBA-Team            & K. Bryant, team, LA Lakers        & $41/164$ & $9$ \\ 
& Oscars              & Gravity, director, A. Cuar\'{o}n  & $78/4680$ & $13$ \\
& CEO*                & Best Buy, keyPerson, H. Joly      & $201/1208$ & $107$ \\
& US War*             & First Battle of Bull Run, battle, I. McDowell & $126/710$ & $150$ \\
& US-V. President*    & B.~Obama, vicePresident, J. Biden & $47/274$ & $169$ \\
& FLOTUS              & B. Obama, spouse, M. Obama        & $16/256$ & $298$ \\
& US-Capital \#2*         & Alabama, capital, Montgomery        & $50/300$ & $4214$ \\
\midrule
\multirow{6}{*}{Real-World} &
GREC-Birthplace     & D. Snow, birthPlace, Windermere, CA   & $273/1092$ & $8$ \\
& GREC-Deathplace     & N. Tate, deathPlace, Southwark        & $126/504$ & $8$ \\
& GREC-Education      & J. Warga, education, Bach. of Science & $466/1861$ & $9$ \\
& GREC-Institution    & A. Mirsky, almaMater, Harvard College & $1546/6184$ & $11$ \\
& WSDM-Nationality    & A. Einstein, nationality, Germany     & $50/200$ & $97$ \\
& WSDM-Profession     & A. Sandler, profession, Comedian      & $110/440$ & $220$ \\
\bottomrule
\end{tabular}
\end{table*}

\subsubsection{Benchmark \& Metric}

We compare our approaches to three existing algorithms designed for fact checking (Knowledge Linker~\cite{ciampaglia2015computational}, PredPath~\cite{shi2016discriminative}, and PRA~\cite{lao2010relational}), one algorithm for knowledge graph completion (TransE~\cite{bordes2013translating}), and four link prediction algorithms (Katz~\cite{katz1953new}, Adamic \& Adar~\cite{adamic2003friends}, Jaccard coefficient~\cite{liben2007link}, and Degree Product~\cite{shi2016discriminative}). We use the area under the Receiver Operating Characteristic curve (AUROC) as a metric to evaluate algorithms; it allows us to compare the accuracy across datasets with different ratios of true and false facts. Each method emits a list of probabilistic scores, one for each triple, and the AUROC expresses the probability that a true triple receives a higher score than a false one. 

\subsubsection{Implementation \& Configuration}

All algorithms have been implemented in Python~2.7, and we use Cython~0.22 to efficiently compute single-source shortest paths and distances as required for KS (\prettyref{algline:pathfinding} and~\ref{algline:pathfinding2} in \prettyref{alg:knowledge-stream}). The source code for our methods can be found at \url{https://github.com/shiralkarprashant/knowledgestream}. For Katz, PRA and PredPath, we use up to 200 paths for every value of path length $l = 1, 2, 3$. In the case of TransE, we create 100-dimensional embeddings using a margin of one and a learning rate of 0.01 for 1,000 epochs.

\subsection{Results}

\subsubsection{Fact checking}
\label{sec:results}

\begin{table*}[tbp]
\renewcommand{\arraystretch}{1.3}
\caption{Fact-checking performance (AUROC) on synthetic data. Best scores for each dataset are shown in bold.}
\label{tab:fact-checking-results-synthetic}
\centering 
\begin{tabular}{lccccccccc}
\toprule
\textbf{Method} 
& \textbf{NYT-Bestseller} 
& \textbf{NBA-Team}
& \textbf{Oscars}
& \textbf{CEO}
& \textbf{US-War}
& \textbf{US-V. President}
& \textbf{FLOTUS}
& \textbf{Capital \#2} 
& \textbf{Avg. (S.E.)}  \\ 
\midrule
KL-REL  & 96.32 & 99.94 & 97.67 & \textbf{89.88} & 86.34 & 87.29 & 98.32 & \textbf{100.00} & 94.47 (2.0) \\
KS      & 89.72 & \textbf{99.96} & 95.00 & 81.19 & 72.11 & 77.80 & 98.05 & \textbf{100.00} & 89.23 (3.9) \\
KS-AVG  & 91.95 & 99.01 & 98.13 & 80.96 & \textbf{99.98} & \textbf{99.53} & 99.09 & 99.76 & 96.05 (2.3) \\
KS-CV   & 93.63 & 99.29 & 97.72 & 80.52 & \textbf{99.98} & 99.47 & 99.27 & 99.28 & 96.14 (2.3) \\ 
\midrule
PredPath~\cite{shi2016discriminative}   & \textbf{99.80} & 92.31 & \textbf{99.97} & 88.67 & 99.51 & 94.40 & \textbf{100.00} & 99.68 & \textbf{96.79} (1.6) \\
KL~\cite{ciampaglia2015computational}   & 94.99 & 99.94 & 97.56 & 89.77 & 63.55 & 74.62 & 98.59 & 99.42 & 89.80 (4.8) \\
PRA~\cite{lao2010relational}            & 96.24 & 91.26 & 99.54 & 87.73 & 99.96 & 50.00 & 60.48 & 98.88 & 85.51 (6.8) \\
TransE~\cite{bordes2013translating}     & 80.99 & 56.71 & 82.66 & 82.68 & 53.22 & 72.50 & 84.82 & 85.31 & 74.86 (4.6) \\
Katz~\cite{katz1953new}                   & 96.52 & 98.50 & 98.98 & 87.53 & 57.80 & 72.92 & 97.42 & 99.97 & 88.70 (5.5) \\
Adamic \& Adar~\cite{adamic2003friends}             & 95.84 & 99.73 & 56.54 & 84.97 & 54.98 & 81.06 & 99.40 & \textbf{100.00} & 84.06 (6.7) \\
Jaccard~\cite{liben2007link}                 & 92.64 & 99.42 & 53.35 & 78.74 & 49.68 & 70.79 & 97.89 & \textbf{100.00} & 80.31 (7.3) \\
Degree Product~\cite{shi2016discriminative}         & 56.52 & 53.21 & 54.42 & 49.17 & 64.08 & 49.55 & 50.00 & 52.10 & 53.63 (1.7) \\
\bottomrule
\end{tabular}
\end{table*}

\prettyref{tab:fact-checking-results-synthetic} and \prettyref{tab:fact-checking-results-real} give a comparison of fact-checking performance between our approaches and other algorithms on several synthetic and real-world datasets. We report average performance and standard error across datasets for each method in the last column. Although statistical significance tests do not reveal a clear overall winner, we can make a few observations. 
KL-REL performs better than the original KL, TransE, and all link prediction algorithms. In fact, it outperforms all other algorithms on real-world datasets and has comparable performance to PredPath on synthetic data. 

\begin{figure}[tbp]
    \centering
    \includegraphics[scale=0.5]{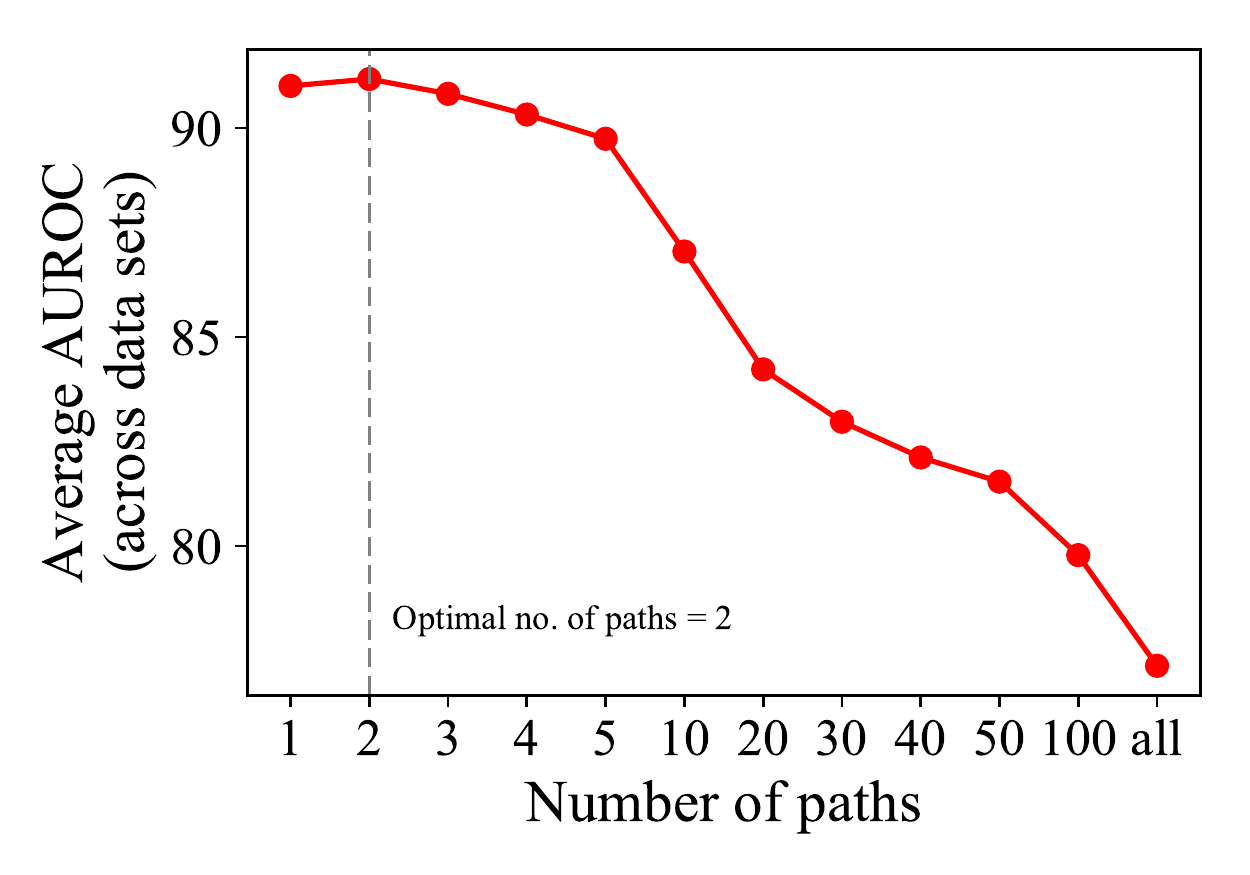}
    \vspace{-1em}
    \caption{Average performance of Knowledge Stream across datasets as a function of the number of paths used in the stream.}
    \label{fig:avg-KS-performance}
\end{figure}

KS lags behind KL. A possible explanation could be that the extra signal provided by the additional paths found by KS may not always be beneficial. To shed more light into this issue, we analyzed the average performance as a function of the number of paths in the stream.  \prettyref{fig:avg-KS-performance} shows that the overall optimum is attained when exactly two paths are considered. On the one hand, this confirms the value of considering multiple paths. On the other hand, this suggests that too many paths hinder performance, and thus the number of paths should be tuned.

Based on this insight, we include in our evaluation two variants of Knowledge Stream. KS-AVG uses the number of paths (two) resulting in the best performance on average. KS-CV uses cross-validation to tune the optimal number of paths for each dataset; this makes KS-CV a supervised approach. As we see from both tables, KS-AVG and KS-CV have a better performance on average than KS, and even better than KL-REL on synthetic datasets. This confirms our intuition of focusing only on a few paths in a stream.

We observe that our algorithms (KL-REL, KS, KS-AVG and KS-CV) often outperform existing fact-checking methods (PredPath, KL and PRA). We emphasize that KL-REL and KS-AVG are purely unsupervised algorithms, whereas PredPath and PRA require supervision for both feature selection and model training. 

Finally, link prediction algorithms (Adamic \& Adar, Jaccard coefficient and Degree Product) tend to perform poorly. Katz is the exception in this category. On real-world datasets, its performance is comparable to that of KL-REL. However, both KS or KL-REL are computationally efficient compared to Katz. In the case of KL-REL this is because of its focus on a single path. As for KS, it uses multiple paths and penalizes longer paths just like Katz, but is more efficient thanks to the capacity constraints.


\begin{table*}[tbp]
\renewcommand{\arraystretch}{1.3}
\caption{Fact-checking performance (AUROC) on real test datasets. Best scores for each dataset are shown in bold.}
\label{tab:fact-checking-results-real}
\centering 
\begin{tabular}{lC{0.16\columnwidth}C{0.16\columnwidth}C{0.16\columnwidth}C{0.16\columnwidth}C{0.16\columnwidth}C{0.16\columnwidth}C{0.16\columnwidth} }
\toprule
\textbf{Method} 
& \textbf{GREC Birthplace} 
& \textbf{GREC Deathplace} 
& \textbf{GREC Education} 
& \textbf{GREC Institution} 
& \textbf{WSDM Nationality} 
& \textbf{WSDM Profession} 
& \textbf{Avg.\newline (S. E.)} \\ 
\midrule
KL-REL  & \textbf{92.54} & \textbf{90.91} & 86.44 & 85.64 & 96.92 & 97.32 & \textbf{91.63} (2.0) \\
KS      & 72.92 & 80.02 & 89.03 & 78.62 & 97.92 & 98.66 & 86.20 (4.4) \\
KS-AVG  & 81.38 & 83.58 & 75.46 & 81.31 & 93.37 & 92.93 & 84.67 (2.9) \\
KS-CV   & 82.28 & 82.57 & 75.23 & 81.33 & 94.20 & 95.84 & 85.24 (3.3) \\ 
\midrule
PredPath~\cite{shi2016discriminative}   & 84.64 & 76.54 & 83.21 & 80.14 & 95.20 & 92.71 & 85.41 (2.9) \\
KL~\cite{ciampaglia2015computational}   & 92.10 & 90.49 & 62.32 & \textbf{87.61} & 96.05 & 91.36 & 86.65 (5.0) \\
PRA~\cite{lao2010relational}            & 74.34 & 75.58 & 70.51 & 63.95 & 83.87 & 50.00 & 69.71 (4.8) \\
TransE~\cite{bordes2013translating}     & 54.88 & 56.47 & 66.32 & 44.99 & 77.09 & 82.91 & 63.78 (5.9) \\
Katz~\cite{katz1953new}                   & 88.46 & 84.07 & 89.55 & 82.99 & \textbf{99.23} & \textbf{98.84} & 90.52 (2.9) \\
Adamic \& Adar~\cite{adamic2003friends}             & 82.79 & 79.13 & 50.00 & 74.58 & 97.21 & 95.07 & 79.80 (7.0) \\
Jaccard~\cite{liben2007link}                 & 80.39 & 75.99 & 49.95 & 69.88 & 95.93 & 90.01 & 77.02 (6.6) \\
Degree Product~\cite{shi2016discriminative}         & 52.82 & 50.86 & \textbf{91.51} & 64.56 & 84.38 & 86.36 & 71.75 (7.3) \\
\bottomrule
\end{tabular}
\end{table*}

\subsubsection{Discovery of relational patterns}
\label{sec:relational-pattern-discovery}

For each triple, the paths discovered by algorithms like KS, KL-REL, PRA, and PredPath can be seen as the evidence used by the algorithm in deciding whether the fact is true. By pooling together evidence from many triples, we can discover data-driven patterns that define a relation, based on the prior knowledge in the KG. 
It is natural to ask whether the patterns discovered by our methods conform to common-sense understanding of these relations. To do so, we perform the following simple exercise. For each relation, we define the two sets $A$ and $B$ of all paths discovered from either true or false triples, respectively. We then rank the  paths in decreasing order of their frequency of occurrence in the set difference $A-B$. 

\prettyref{tab:relational-patterns} shows a few top patterns discovered by KS for a few relations. The patterns are highly relevant. We omit many other interesting examples due to space constraints.
This characteristic of KS has a wider applicability --- with only a few true and false examples, these patterns can be discovered in an unsupervised fashion, and used either as a seed set of rules in information extraction projects, or as features for learning other concepts. 

\begin{table*}[!t]
\renewcommand{\arraystretch}{1.3}
\caption{Relational patterns discovered by Knowledge Stream.}
\label{tab:relational-patterns}
\centering 
\begin{tabular}{ m{0.06\columnwidth}|C{0.5\columnwidth}|R{0.09\columnwidth}|L{1.15\columnwidth} }
\toprule
\multicolumn{1}{c}{\textbf{Relation}} & \multicolumn{1}{|c}{\textbf{Pattern}} & \multicolumn{1}{|c}{\textbf{Freq.}} & \multicolumn{1}{|c}{\textbf{Example}} \\
\midrule
\multirow{4}{4em}{\begin{sideways}Spouse \,\end{sideways}} & (child, \noarrowinvrelation{child}) & 34 & J. F. Kennedy $\relation{child}$ Patrick Kennedy $\invrelation{child}$ Jacqueline Kennedy Onassis \\ 
& (\noarrowinvrelation{parent}, parent) & 20 & J. F. Kennedy $\invrelation{parent}$ Patrick Kennedy $\relation{parent}$ Jacqueline Kennedy Onassis \\
& (child, parent) & 19  & J. F. Kennedy $\relation{child}$ Patrick Kennedy $\relation{parent}$ Jacqueline Kennedy Onassis \\
& (predecessor, spouse, \noarrowinvrelation{predecessor}) & 6 & R. Reagan $\relation{predecessor}$ P. Brown $\relation{spouse}$ B. Brown $\invrelation{predecessor}$ N. Reagan \\ 
\midrule
\multirow{4}{4em}{\begin{sideways}CEO \,\end{sideways}} & (\noarrowinvrelation{parentCompany}, keyPerson) & 32 & News Corporation $\invrelation{parentCompany}$ Sky TV plc $\relation{keyPerson}$ Rupert Murdoch \\
& (\noarrowinvrelation{employer}) & 24 & Twitter $\invrelation{employer}$ Dick Costolo \\
& (foundedBy) & 24 & Foxconn $\relation{foundedBy}$ Terry Gou \\
& (subsidiary, keyPerson) & 20 & Samsung $\relation{subsidiary}$ Samsung Electronics $\relation{keyPerson}$ Lee Kun-hee \\ 
\midrule
\multirow{4}{4em}{\begin{sideways}US-Capital \,\end{sideways}} & (\noarrowinvrelation{deathPlace}, deathPlace) & 491 & Delaware $\invrelation{deathPlace}$ Nathaniel B. Smithers $\relation{deathPlace}$ Dover, Delaware \\
& (part, isPartOf) & 123 & Delaware $\relation{part}$ Delaware Valley $\relation{isPartOf}$ Dover, Delaware \\
& (\noarrowinvrelation{headquarter}, location) & 112 & Kansas $\invrelation{headquarter}$ State Library of Kansas $\relation{location}$ Topeka, Kansas \\
& (\noarrowinvrelation{jurisdiction}, location) & 104 & Kansas $\invrelation{jurisdiction}$ Kansas Department of Revenue $\relation{location}$ Topeka, Kansas \\
\bottomrule
\end{tabular}
\vspace{-1em}
\end{table*}

\subsubsection{Surfacing facts relevant to a target claim}
\label{sec:relational-facts-discovery}

\begin{figure*}[tbp]
    \centering
    \includegraphics[width=\textwidth]{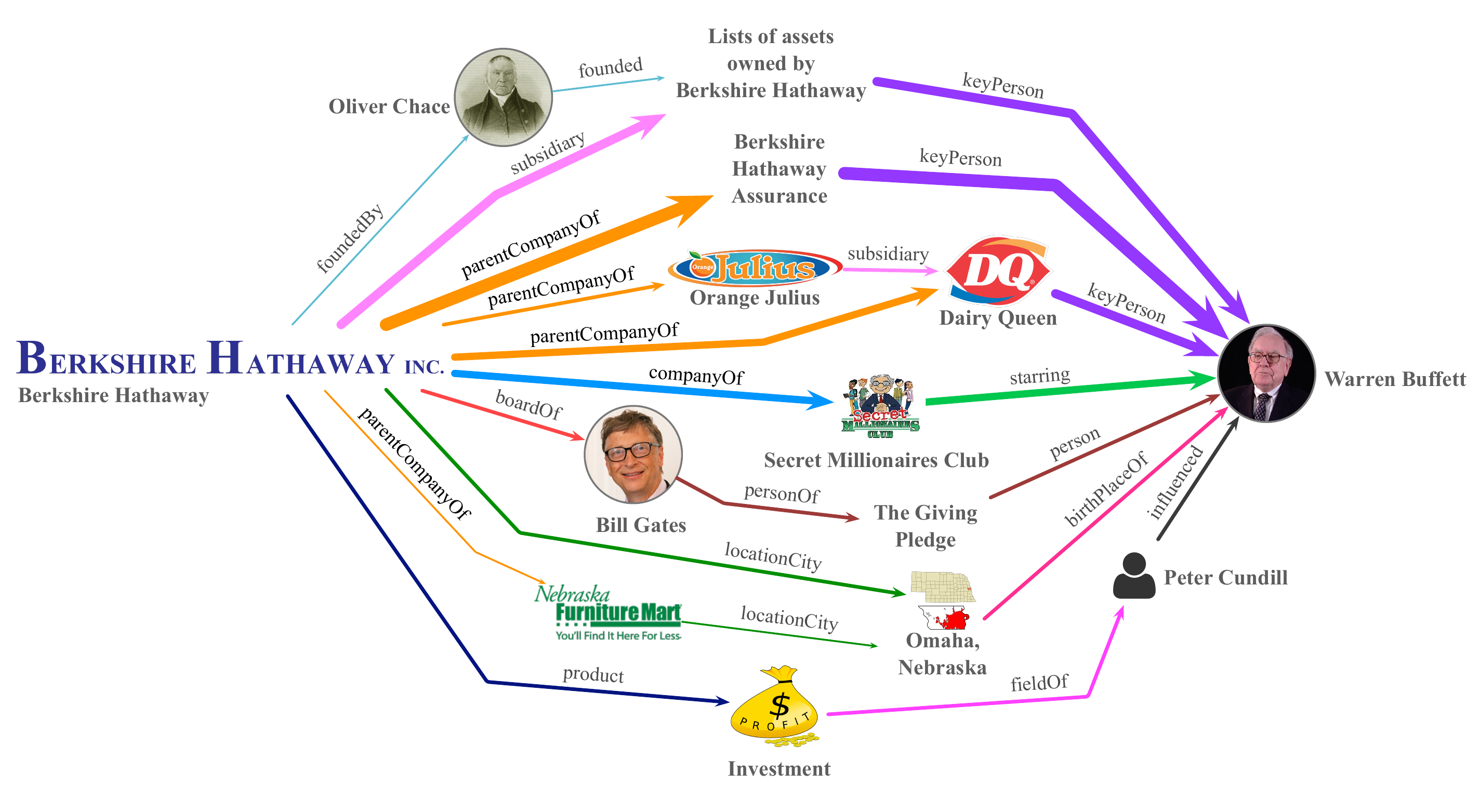}
    \vspace{-1em}
    \caption{Relevant facts about a target claim as surfaced by Knowledge Stream.}
    \label{fig:relevant-facts}
\end{figure*}

The workflow of a human fact checker begins by gathering facts that are relevant to the claim being checked. Possible sources are background information, interview transcripts, etc.~\cite{borel2016chicago}. We find that KS can assist in this task by identifying the general context of a triple. As an illustration, \prettyref{fig:relevant-facts} shows the set of most relevant facts (as indicated by the paths) for the triple {\tt (Berkshire Hathaway, keyPerson, Warren Buffett)}, with the width of edges roughly proportional to their net flow~$\pathweight[]$. See \prettyref{fig:ks-example} for another example. Notice the diversity in the set of facts that support these triples. Also note how Knowledge Stream is able to ``bubble up'' the most intuitively relevant facts by channeling a large flow through their corresponding paths (indicated by their wider edges). Other approaches rely on the availability of path patterns that are either curated by knowledge engineers or mined using a large number of labeled examples. KS automatically surfaces relevant ground facts in an unsupervised way. We believe that it is the first computational fact-checking approach  featuring such an expressive power.  

\section{Related Work}
\label{sec:related-work}

Fact checking is an important activity to prevent dubious claims and unverified rumors from spreading. Preliminary computational approaches have employed metadata and other contextual indicators around entities of interest, e.g., characteristic features in user account metadata, unexpected shifts in temporal signals, credibility, and so on. For example, Truthy~\cite{ratkiewicz2011truthy}, Rumorlens~\cite{resnick2014rumorlens}, TweetCred~\cite{Gupta2014}, and ClaimBuster~\cite{hassan2015quest} are systems whose aim is to study the spread of misinformation and rumors, and identify interesting claims to check. The Hoaxy system~\cite{shao2016hoaxy} tracks claims and fact checks to study their interplay. By design, these systems do not attempt to understand the actual contents of claims, which limits their applicability.

Other approaches focus on checking the content of a claim based on prior knowledge, typically found in a knowledge base or knowledge graph. We can distinguish two broad classes of methods based on how easy it is to interpret their results. On the one hand, we have approaches inspired by logical reasoning (e.g., ILP~\cite{muggleton1992inductive} and AMIE~\cite{galarraga2013amie}), which mine  first-order Horn clauses and are thus easy to interpret. On the other, there are statistical learning models (e.g., RESCAL~\cite{nickel2016review}, TransE~\cite{bordes2013translating}, TransH~\cite{wang2014knowledge}, TransR~\cite{lin2015learning}, and ProjE~\cite{Shi2017ProjEEP}) that create vector embeddings for entities and relations, which can be used to assign similarity scores. Statistical approaches can be particularly hard to interpret, but they are great at handling uncertainties and capturing global regularities in the KG. Unfortunately, scalability is an issue for both types of approaches, as many of the algorithms mentioned above struggle to perform in the face of large-scale KGs,  due to either large search spaces or high model complexity. Nickel \textit{et al.}~\cite{nickel2016review} review a number of these models.

Only a few approaches fall somewhere in the middle of this interpretability spectrum. Ciampaglia \textit{et al.}~\cite{ciampaglia2015computational} propose an approach that relies on a single short, specific path to differentiate a true fact from a false one. Although intuitive, their algorithm fails to account for the semantics of the target predicate.

PRA~\cite{lao2010relational} and PredPath~\cite{shi2016discriminative} mine the KG in search of paths connecting the subject to the object of a triple, and use the predicate labels found along these paths to identify features for a supervised learning framework. Labeled examples of true and false triples are therefore needed at the stage of feature selection and during model training. Both approaches spend significant computational resources on feature generation and selection. And even though they rely on massive amounts of features, most provide only a very weak signal. Nevertheless, they have been shown to be very effective on fact-checking test cases~\cite{shi2016discriminative}, and in large-scale machine reading projects~\cite{lao2011random,dong2014knowledge}. They also offer some interpretability due to the features (or rules) they learn. Our methods achieve comparable or better performance while offering greater interpretability and expressiveness in terms of supporting facts, and without training --- except for ``learning'' the edge capacities (see \prettyref{sec:relational-similarity}). 

Most of the approaches (including ours) described above have focused on checking claims as simple as a triple. Since a triple is a link in the graph, an impressive array of link prediction algorithms in dynamic networks~\cite{maguitman2006algorithmic, liben2007link, lu2011link} can be applied to the task of fact checking. However, these approaches mainly rely on elementary structural cues, leading to poor performance on many fact-checking test sets~\cite{shi2016discriminative} (see also \prettyref{tab:fact-checking-results-real}).

\section{Discussion and future work}
\label{sec:conclusion-futurework}

Network flow theory~\cite{ahuja1993network} has guided the design of many applications in engineering, logistics, manufacturing, and so on. In this paper, we have shown that it can also serve as a useful toolbox for reasoning about facts, and for fact checking in particular.

We presented two novel, unsupervised approaches to assess and explain the truthfulness of a statement of fact by leveraging its semantic context in a knowledge graph. Knowledge Stream is based on network flow and employs multiple short paths; Relational Knowledge Linker finds a single shortest path. In pursuing both approaches, we also proposed a novel method to measure the similarity of any two relations purely based on their co-occurrence in the KG.

We evaluated both approaches on a diverse set of real-world and synthetic test cases, and found that their performance is on par with the state of the art. Moreover, we saw that, in many cases, multiple paths can provide additional evidence to support fact checking. Our Knowledge Stream model offers high expressive power by its ability to automatically surface useful path patterns and relevant facts about a claim. Based on this experience we believe that network flow techniques are particularly promising for fact checking.

Knowledge Stream is still a preliminary approach for computational fact checking, leaving much room for improvement to address complex test cases. For example, the success of KS hinges on the appropriate design of edge capacities in the graph. We have explored the use of relational similarity for this purpose. The development and evaluation of effective relational similarity metrics is an important avenue of future work. The capacities could also incorporate metadata from the KG itself, for example confidence scores from the information extraction phase (see, e.g., YAGO~\cite{Suchanek:2007:YCS:1242572.1242667}).

In surfacing facts relevant to a triple, we ranked the set of paths in a stream based on their flow values. Devising alternative ways to rank such facts, reflecting their novelty, diversity, or serendipity (as is done in evaluating recommender systems) is another interesting thread of future research.

Many KGs (e.g., YAGO2~\cite{hoffart2013yago2} and Wikidata~\cite{erxleben2014introducing}) now contain facts augmented with spatio-temporal details. Checking facts that may be true only during a certain time frame or at a certain location is another important challenge. One way to extend KS to handle such facts could be to bias the search toward those areas of the KG that may contain facts valid during that interval or near that place. 

Lastly, our version of KS relies on successive path-finding, which can be slow for triples involving subjects with a large search space. Our implementation takes a few minutes to check each triple with DBpedia. Other approaches could be explored in the future. For example, the network simplex algorithm~\cite{ahuja1993network} has better theoretical and run-time behavior.

\section*{Acknowledgments}

The authors would like to thank B. Shi and T. Weninger for sharing their evaluation data. This work was supported in part by NSF (Award CCF-1101743) and DARPA (grant W911NF-12-1-0037). Funding agencies had no role in study design, data collection and analysis, decision to publish, or preparation of the manuscript. 

\bibliographystyle{IEEEtran}
\bibliography{bibliography}

\end{document}